\documentclass[journal]{IEEEtran}
\ifCLASSINFOpdf
\else
\fi
\usepackage{graphicx}
\usepackage{multirow} 
\usepackage{makecell}
\usepackage{booktabs}
\usepackage{amsfonts}
\usepackage{amsmath}
\usepackage{ulem}
\usepackage{hyperref}
\usepackage{float}
\usepackage{marvosym}
\usepackage{algorithm}
\usepackage{algorithmic}
\usepackage{subcaption}

\def\etal{\normalem\emph{et al.~}}

\def\aka{\normalem\emph{a.k.a.}}

\begin{document}
%
\title{MatchSeg: Towards Better Segmentation \\ via Reference Image Matching}
%
%
%

\author{
Jiayu Huo$^{1,\#,*}$,
Ruiqiang Xiao$^{2,\#}$,
Haotian Zheng$^{3}$,
Yang Liu$^{1}$,
S\'{e}bastien Ourselin$^{1}$,
Rachel Sparks$^{1}$ \\

$^{1}$ School of Biomedical Engineering and Imaging Sciences (BMEIS), King's College London, London, UK. \\
$^{2}$ School of Science, The Hong Kong University of Science and Technology, Hong Kong, China. \\
$^{3}$ Sydney Smart Technology College, Northeastern University, Shenyang, China. \\
$^{\#}$Equal Contribution \\
$^{*}$Correspnding author: jiayu.huo@kcl.ac.uk 

}

\maketitle

\begin{abstract}
Recently, automated medical image segmentation methods based on deep learning have achieved great success. However, they heavily rely on large annotated datasets, which are costly and time-consuming to acquire. Few-shot learning aims to overcome the need for annotated data by using a small labeled dataset, known as a support set, to guide predicting labels for new, unlabeled images, known as the query set. Inspired by this paradigm, we introduce MatchSeg, a novel framework that enhances medical image segmentation through strategic reference image matching. We leverage contrastive language-image pre-training (CLIP) to select highly relevant samples when defining the support set. Additionally, we design a joint attention module to strengthen the interaction between support and query features, facilitating a more effective knowledge transfer between support and query sets. We validated our method across four public datasets. Experimental results demonstrate superior segmentation performance and powerful domain generalization ability of MatchSeg against existing methods for domain-specific and cross-domain segmentation tasks. Our code is made available at \url{https://github.com/keeplearning-again/MatchSeg}
\end{abstract}


%
\IEEEpeerreviewmaketitle

\section{Introduction}
\IEEEPARstart{M}{edical} image segmentation plays a crucial role in computer-aided diagnosis as it can provide quantitative information on regions of interest such as brain tumors and skin lesions~\cite{al2024machine,jiang2018medical}. The advent of deep learning has significantly facilitated image segmentation algorithms, with deep learning methods achieving state-of-the-art performance for both accuracy and efficiency in segmenting anomalies and anatomical structures~\cite{conze2023current}. Generally, deep learning algorithms follow a fully supervised learning paradigm, necessitating large annotated datasets prepared by domain experts, which is time-consuming and costly~\cite{tajbakhsh2020embracing,jiao2023learning}. The dependency on a large number of labels for supervised learning has resulted in several self-supervised and semi-supervised approaches being developed to reduce the need for annotations~\cite{ouyang2022self,you2024rethinking}. However, the performances of these methods are often constrained by the availability and quality of unlabeled data, both of which may be sparse in the medical domain. Furthermore, the generalization ability of semi-supervised and self-supervised learning frameworks is limited, making it difficult to rapidly transfer models to new domains~\cite{li2021generative,you2024rethinking,koehler2022noisy}.

Few-Shot Learning (FSL)~\cite{li2021adaptive,wang2019panet,butoi2023universeg} emerges as a compelling solution, adept at navigating the constraints of scarce data by leveraging a minimal set of labeled examples to make accurate predictions on novel, unseen classes. This approach depends on the concept of learning-to-learn, where a meta-learner is trained to generalize from a small support set of samples with labels to effectively segment query images without further retraining. Li \etal\cite{li2021adaptive} designed an adaptive prototype learning method via the superpixel technique for the object segmentation task. Wang \etal\cite{wang2019panet} proposed a bi-directional prototype learning strategy to learn the representative feature embeddings from query and support features for few-shot segmentation. The success of the few-shot paradigm lies in how effectively the knowledge can be transferred from the support set to query images. Inspired by such a design, Butoi \etal\cite{butoi2023universeg} designed a reference image segmentation called UniverSeg which predicts a mask for a region of interest (ROI) in the query image through guidance provided by the support image-mask pairs. However, it is limited by the random choice of the support set which results in a large variance in the model's performance dependent on the distribution of samples in the support set.

We introduce MatchSeg, a novel framework that enhances medical image segmentation through strategic reference image matching. Distinct from conventional FSL approaches, MatchSeg leverages guidance from the contrastive language-image pre-training \aka~CLIP~\cite{radford2021learning} image encoder to select a highly relevant support set, thereby facilitating more accurate and robust segmentation results. Furthermore, we propose a joint attention module that learns deeper connections between query and support features, enabling a more seamless and effective knowledge transfer. Our comprehensive evaluation, conducted across four public datasets, demonstrates the superiority of MatchSeg against other deep learning algorithms including SOTA full-supervised and few-shot learning models. Additionally, we validate the effectiveness of each component within our framework to demonstrate the effectiveness of our approach. 

\begin{figure*}[!htbp]
\centering
\includegraphics[width=\linewidth]{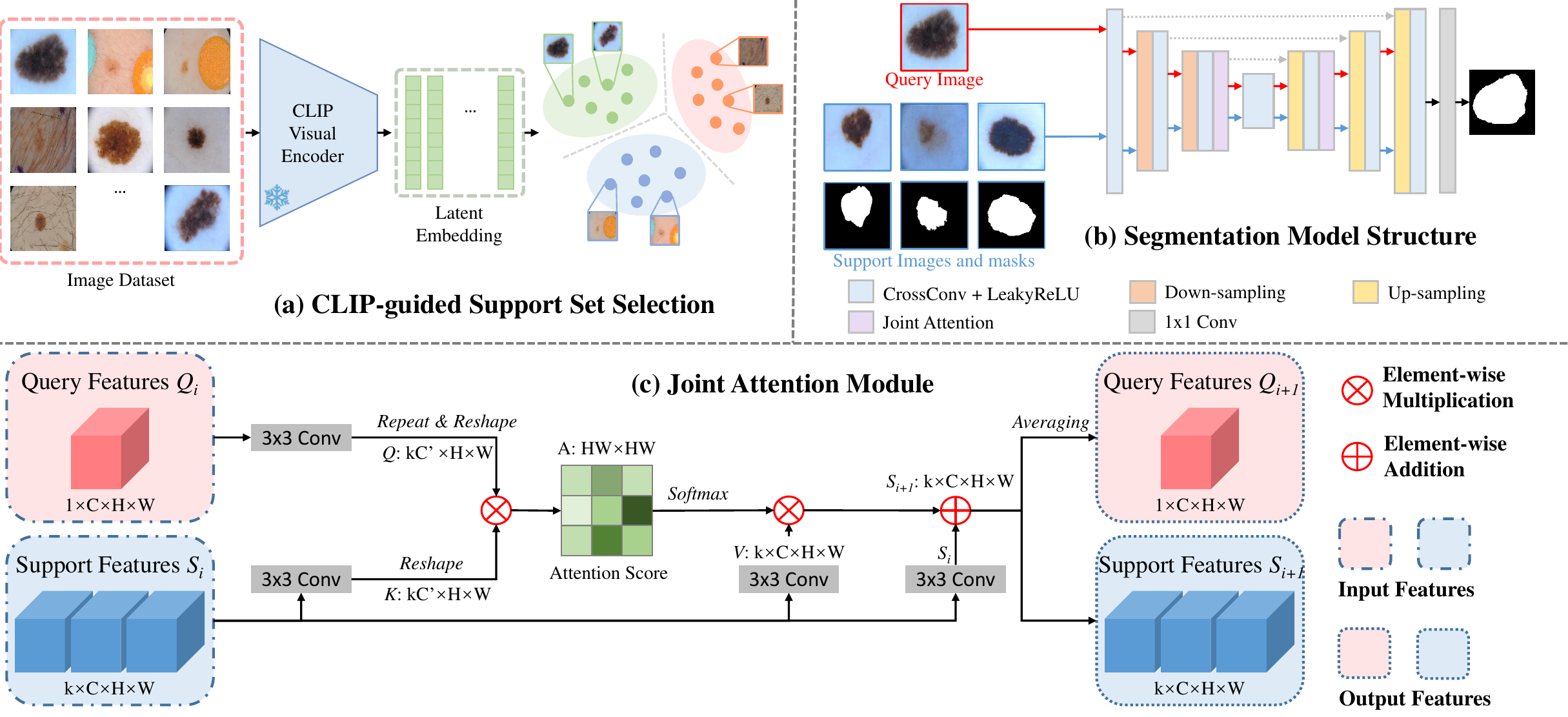}
\caption{The MatchSeg framework comprises (a) support set selection using a CLIP image encoder, (b) the segmentation model, and (c) the Joint Attention module. In the Joint Attention module, $k$ is the number of support image-label pairs. $C$ is the number of channels of the feature map. $C'$ is the reduced number of channels for lower computational cost. $H\times W$ is the spatial feature size.}
\label{fig:main_fig}
\end{figure*}

\section{Methodology}
Fig.~\ref{fig:main_fig} illustrates the MatchSeg framework for image segmentation. Specifically, MatchSeg contains two steps: CLIP-guided support set selection (Fig.~\ref{fig:main_fig}a) and segmentation model training (Fig.~\ref{fig:main_fig}b). To enhance the feature interaction between the query image and the support set we present the Joint Attention module, as shown in Fig.~\ref{fig:main_fig}c. We provide a detailed description of each component in the following subsections. 

\subsection{CLIP-guided Support Set Selection}
\label{sec:CLIP}
The major challenge MatchSeg aims to address is to select the most effective support set to give stable and accurate segmentation results. The hypothesis underpinning our work is that an effective support set is highly correlated with the possible query images, and by selecting support sets based on their relationship with the query image the segmentation is most likely to be accurate. Given the capabilities of foundational models in semantic interpretation, we decided to leverage the CLIP image encoder as the cornerstone for global feature extraction to obtain a support set that is similar to the query image. The flexibility of the CLIP visual encoder in aligning language and vision semantics positions makes it a formidable tool for defining features to compare image appearance similarities.

Fig.~\ref{fig:main_fig}a shows the process of using the CLIP image encoder to find the suitable support set. Specifically, we first input all potential support set images and the query image into the CLIP image encoder to get a latent representation for each image. Then cosine similarities between the query image embedding and embeddings of the potential support set images are computed. The potential support set images are ranked in descending order. We then select the top $K$ image-label pairs to form the support set. The pseudo-code of this step is described in Algorithm~\ref{alg:CLIP_match}.

\begin{algorithm*}[!t] 
\caption{Retrieve top-$K$ most similar support image-label pairs using the CLIP image encoder}
\begin{algorithmic}[1]
\REQUIRE Pre-trained CLIP image encoder $f_\theta$; Query image $x_q$; Support dataset $\mathcal{S}$; Support set size $K$
\ENSURE top-$K$ support set $S_k$
\STATE Cosine similarity list $\mathbb{C}$ = empty list
\STATE Query feature $z_q$ = $f_\theta$($x_q$)
\FOR{index $i$, support image $x_s$ in $\mathcal{S}$}
    \STATE Support feature $z_s$ = $f_\theta$($x_s$)
    \STATE Calculate cosine similarity value $c$ = $\cos$($z_q$,$z_s$)
    \STATE Add \{$c$,$i$\} to $\mathbb{C}$
\ENDFOR
\STATE Sort $\mathbb{C}$ in descending order by $c$
\STATE Select indices of the top-$K$ entries from $\mathbb{C}$
\STATE Retrieve the corresponding images $X_s$ and labels $Y_s$ from $\mathcal{S}$ using the selected indices
\STATE Add \{$X_s$,$Y_s$\} to $S_k$
\RETURN $S_k$
\end{algorithmic}
\label{alg:CLIP_match}
\end{algorithm*}

\subsection{Joint Attention Module}
\label{sec:joint}
Fig.~\ref{fig:main_fig}b displays the detailed structure of the segmentation model in MatchSeg. A cross-convolution operator plus a leakyReLU activation function forms a basic block. We utilize bilinear interpolation for both down-sampling and upsampling layers. Furthermore, we introduce a joint attention module for learning an effective interaction between query and support set features in order to improve the ability of the model to capture the global semantic information.

The joint attention module architecture is shown in Fig.~\ref{fig:main_fig}c. For support set features denoted $S_{i}$ and the query features denoted $Q_{i}$, we first pass both features through a convolutional layer to reduce the feature channels. The query features are repeated and reshaped to form the query tensor $Q \in \mathbb{R}^{kC' \times H \times W}$, while the support set features act as both key tensor $K \in \mathbb{R}^{kC' \times H \times W}$ and value tensor $V \in \mathbb{R}^{k \times C \times H \times W}$. Note $C' = C / ratio$ to reduce the computational complexity of this step. This design enables learning a pixel-wise relationship between each support feature and the query feature.

The subsequent step involves flattening the spatial dimensions of $Q^{k}$ and $K$ to $\hat{Q^{k}} \in \mathbb{R}^{HW \times kC' }$ and $\hat{K} \in \mathbb{R}^{HW \times kC' }$, respectively. The attention matrix $A$ is then computed and normalized using the softmax function:
\begin{equation}
A = \text{softmax}(\hat{Q^{k}} \hat{K}^T)
\end{equation}

Next, we multiply the attention matrix $A$ with the value tensor $V$ and add the residual of the support set features to yield the new support set features from the joint-attention module. The updated query set feature is computed by averaging the support set features across channel dimensions and reshaping the tensor to $\mathbb{R}^C \times H \times W$:
\begin{equation}
S_{i+1} = \text{conv}(S_{i}) + A \cdot V
\end{equation}
\begin{equation}
Q_{i+1} = \text{Averaging}(S_{i+1}, dim=1)
\end{equation}
\begin{equation}
\text{Joint Attention}(S_{i}, Q_{i}) = S_{i+1}, Q_{i+1}
\end{equation}

This joint attention module enables the comparison of features between the support and query sets to be refined, enhancing the model's capability to discern and align relevant features across the query and support sets in order to improve model performance. Finally, our segmentation model is trained with a loss function that is defined as a weight-sum of the Dice loss, Binary Cross-Entropy loss, and Focal loss, as follows:
\begin{equation}  
\mathcal{L}_{total} = \lambda_1 \mathcal{L}_{\text{Dice}} + \lambda_2 \mathcal{L}_{\text{BCE}} + \lambda_3 \mathcal{L}_{\text{Focal}},
\end{equation}  
$\lambda_{1}$, $\lambda_{2}$ and $\lambda_{3}$ are weighted factors.

\begin{figure}[!t]  
\centering  
\includegraphics[width=0.5\textwidth]{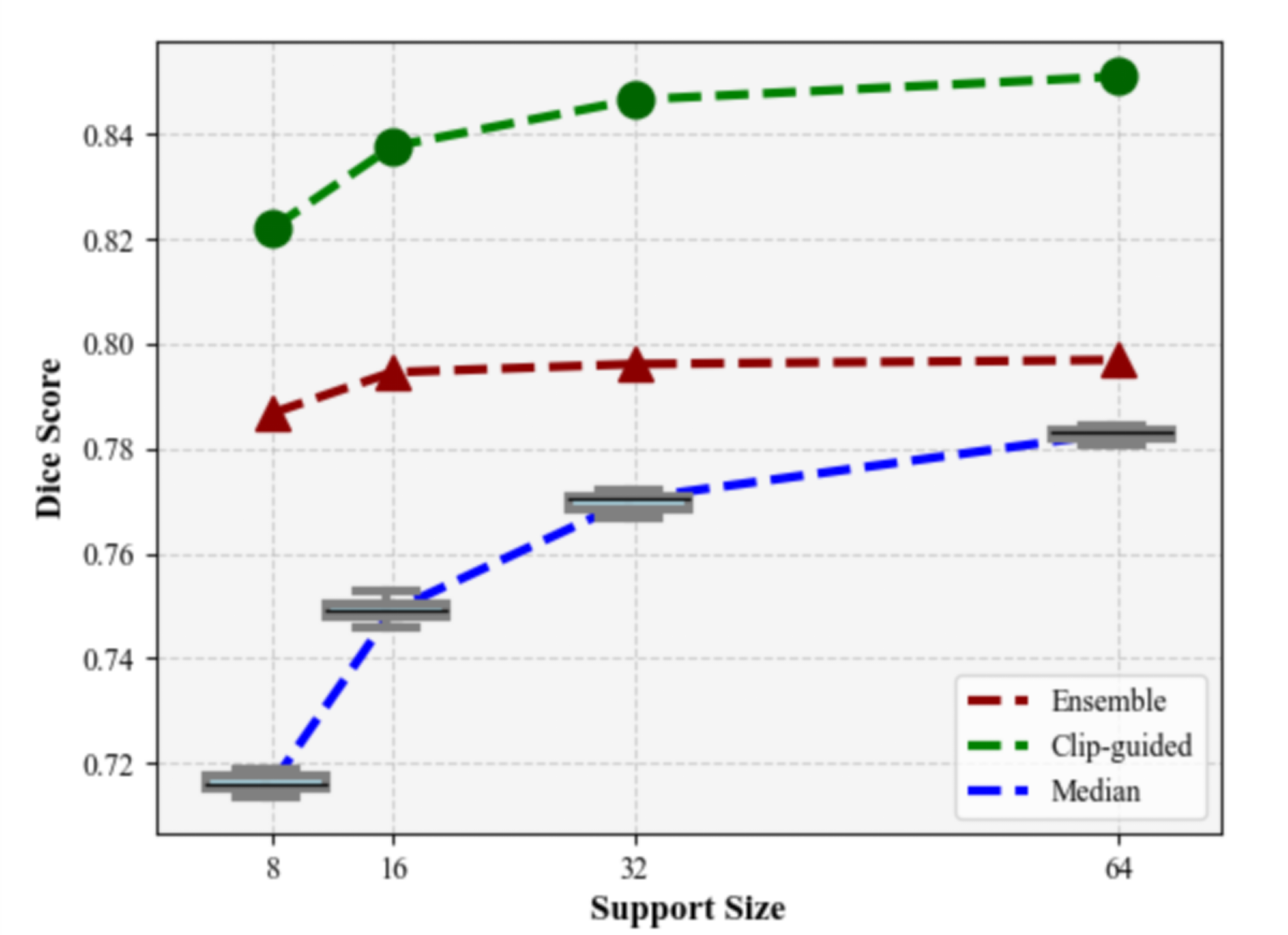}  
\caption{Segmentation performance of UniverSeg with different support set selection strategies and sizes for the HAM10000 dataset.}  
\label{fig:box_plot}  
\end{figure}

\section{Experiments}
\subsection{Datasets}
We evaluated MatchSeg on four public datasets: HAM-10000~\cite{tschandl2018ham10000}, GlaS~\cite{sirinukunwattana2017gland}, BUS~\cite{cheng2010automated}, and BUSI~\cite{al2020dataset}. HAM10000 is a comprehensive dermatoscopic dataset, consisting of 10,015 multi-source images and corresponding masks across 7 lesion types: actinic keratoses and intraepithelial carcinoma (AKIEC), basal cell carcinoma (BCC), benign keratosis-like lesions (BKL), dermatofibroma (DF), melanoma (MEL), melanocytic nevi (NV) and vascular lesions (VASC). GlaS is the colon histology gland segmentation challenge dataset, it has 165 images and masks that have been divided into the training set (85 images) and the test set (80 images) officially. BUS and BUSI are two breast ultrasound datasets that are collected for the breast tumor segmentation task. There are 163 images in the BUS dataset and 647 images in the BUSI dataset. For all datasets, we randomly choose 80\% images as the training set and keep the remaining 20\% as the test set except for the GlaS dataset which is already divided. Note that in the HAM10000 dataset, we split the data in a stratified manner to ensure equal distribution of lesion types between training and testing sets.

\begin{table*}[htbp]  
\centering
\caption{Quantitative results on BUSI, BUS, and GlaS datasets. Best and second-best results are bold and underlined, respectively.}
\fontsize{9}{11}\selectfont
\begin{tabular}{llllllll}  
\toprule
&  & \multicolumn{2}{c}{BUSI} & \multicolumn{2}{c}{BUS} & \multicolumn{2}{c}{GlaS} \\  
\multirow{-2}{*}{Model} &  
  \multirow{-2}{*}{\makecell{Params (M)}} &  
  \multicolumn{1}{c}{DSC$\%\uparrow$} &  
  \multicolumn{1}{c}{IOU$\%\uparrow$} &  
  \multicolumn{1}{c}{DSC$\%\uparrow$} &  
  \multicolumn{1}{c}{IOU$\%\uparrow$} &  
  \multicolumn{1}{c}{DSC$\%\uparrow$} &  
  \multicolumn{1}{c}{IOU$\%\uparrow$} \\ \hline  
UNet~\cite{ronneberger2015u}            &34.53 &79.29             &71.14             &\underline{89.08} &\underline{81.73} &89.58             &82.16 \\  
AttentionUNet~\cite{oktay2018attention} &34.88 &\underline{79.35} &\underline{71.28} &88.12             &80.45             &\underline{90.00} &\underline{82.80} \\
ACC-UNet~\cite{ibtehaz2023acc}          &16.77 &74.59             &65.39             &71.16             &58.98             &77.82             &64.88 \\
\hline
UNext~\cite{valanarasu2022unext}        &1.47  &69.50             &58.80             &70.74             &58.68             &80.96             &69.08 \\  
UCTransNet~\cite{wang2022uctransnet}    &66.34 &78.17             &69.02             &87.96             &79.86             &88.53             &80.15 \\  
\hline
PA-Net~\cite{wang2019panet}             &8.94  &77.02             &67.63             &81.19             &70.45             &84.89             &74.54 \\  
ASG-Net~\cite{li2021adaptive}           &34.06 &50.60             &42.40             &81.14             &70.46             &73.51             &59.28 \\
UniverSeg~\cite{butoi2023universeg}     &1.18  &52.14             &40.21             &59.85             &48.42             &60.79             &44.79 \\
\hline
MatchSeg (ours)                         &7.79  &\textbf{81.03}   &\textbf{72.56}   &\textbf{90.58}   &\textbf{83.23}   &\textbf{90.92}   &\textbf{84.04} \\ 
\bottomrule
\end{tabular} 
\label{tab:in_domain}
\end{table*}

\begin{figure*}[!htb]  
\centering  
\includegraphics[width=0.98\textwidth]{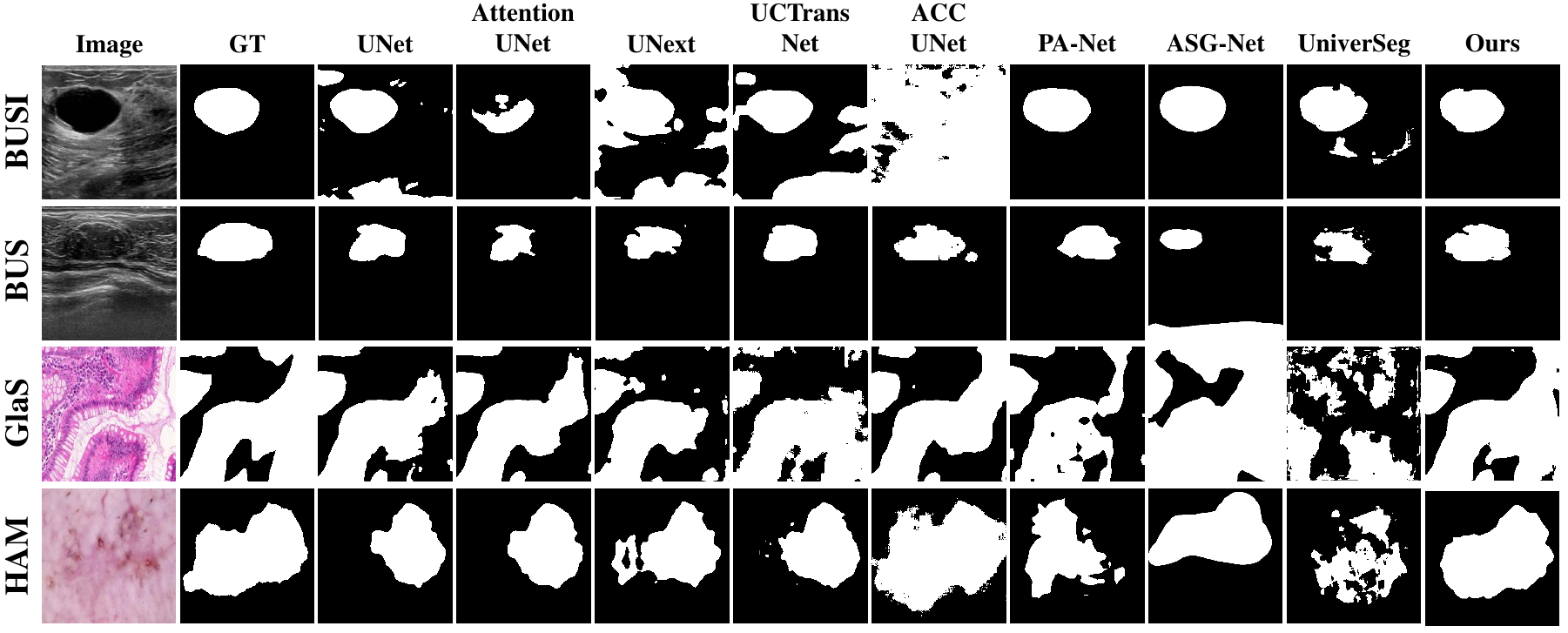}  
\caption{Visualization of different segmentation methods on BUSI, BUS, GlaS and HAM10000 datasets.}  
\label{fig:visualiztion_main}  
\end{figure*} 

\subsection{Experimental Settings}
We design two experiments to evaluate our model, in-domain and cross-domain experiments. For the in-domain experiment, models are trained and evaluated within the same dataset, which allows for an in-depth comparison of our method's capabilities against other segmentation models trained in a fully supervised manner. We utilize three datasets (GlaS, BUI, and BUS) for the in-domain setting. The cross-domain experiment is designed such that we train the model on one domain and test the performance on another domain. This setting is built to analyze the domain generalization ability of the target model. We first interchangeably treat BUSI and BUS as the training and testing dataset. Additionally, in the HAM10000 we designed an out-of-domain experiment where we select a single class, NV, for training MatchSeg and test on all other classes. Given the substantial data imbalance between NV and other classes, a down-sampling operation is implemented on the NV class to ensure the data imbalance does not affect the segmentation performance.

\subsection{Implementation Details}
Our model was implemented on the PyTorch platform, and all experiments were conducted on four NVIDIA GeForce RTX 4090 GPUs. For the support set selection, we utilized the pre-trained ViT-B/16~\cite{dosovitskiy2020image} model that was trained on the LAION-2B~\cite{schuhmann2022laion} dataset. Our model was trained from scratch using the AdamW optimizer initialized with a learning rate of $1e^{-4}$. We set $K$ to 8 to define the support set according to the cosine similarity computed from the CLIP embedding vectors. The input size of all images was resized to $160 \times 160$, with data augmentations including random flipping, random rotation, random scaling, and random cropping. We use the Dice coefficient (DSC) and Intersection over Union (IoU) to evaluate segmentation model performance. The weighted factor $\lambda_{1}$, $\lambda_{2}$ and $\lambda_{3}$ are set to 0.6, 0.3, and 0.3, respectively. These values were determined empirically.

\begin{table*}[!t]
\centering
\caption{Cross-domain segmentation results on BUSI and BUS datasets. A $\rightarrow$ B indicates A for training and B for testing.}
\fontsize{9}{11}\selectfont
\begin{tabular}{llllll}
\toprule
\multirow{2}{*}{Models} &
\multirow{2}{*}{\makecell{Params (M)}} &
\multicolumn{2}{c}{BUSI $\rightarrow$ BUS} &
\multicolumn{2}{c}{BUS $\rightarrow$ BUSI} \\
& &
\multicolumn{1}{c}{DSC$\%\uparrow$} &
\multicolumn{1}{c}{IOU$\%\uparrow$} &
\multicolumn{1}{c}{DSC$\%\uparrow$} &
\multicolumn{1}{c}{IOU$\%\uparrow$} \\ 
\hline  
UNet~\cite{ronneberger2015u}            &34.53  & 74.72             & 67.16             & \underline{56.85} & \underline{49.36} \\
AttentionUNet~\cite{oktay2018attention} &34.88  & \underline{76.18} & \underline{67.66} & \underline{56.85} & 49.32 \\
ACC-UNet~\cite{ibtehaz2023acc}          &16.77  & 61.62             & 53.55             & 42.81             & 33.79 \\
\hline
UNext~\cite{valanarasu2022unext}        &1.47   & 56.37             & 47.30             & 39.93             & 30.92 \\
UCTransNet~\cite{wang2022uctransnet}    &66.34  & 67.52             & 58.21             & 54.91             & 47.16 \\
\hline
PA-Net~\cite{wang2019panet}             &8.94   & 70.99             & 61.80             & 49.15             & 40.28 \\
ASG-Net~\cite{li2021adaptive}           &34.06  & 47.23             & 34.96             & 49.70             & 40.44 \\
UniverSeg~\cite{butoi2023universeg}     &1.18   & 51.75             & 40.91             & 35.19             & 26.05 \\
\hline
MathcSeg (ours)                        &7.79   &\textbf{78.07}     &\textbf{68.86}     &\textbf{59.27}     &\textbf{51.53} \\
\bottomrule
\end{tabular}
\label{tab:bus_cross}
\end{table*}

\begin{table*}[!t]
\centering
\caption{Cross-domain segmentation results on the HAM10000 dataset. Note that all models were trained on the NV lesion type. We use the Dice coefficient to evaluate the segmentation performance.}
\fontsize{9}{11}\selectfont
\begin{tabular}{llccccccc}
\toprule 
Models                                  & Params (M) & AKIEC & BCC & BKL   & DF       & MEL       & VASC       & Average \\  
\hline 
UNet~\cite{ronneberger2015u}             & 34.53  & 75.16             & 68.07             & 84.67             & \underline{83.50} & 91.20             & 81.99             & 83.19 \\
AttentionUNet~\cite{oktay2018attention}  & 34.88  & 76.72             & 69.61             & 85.04             & 83.00             & \underline{92.14} & \textbf{83.71}    & 84.08 \\
ACC-UNet~\cite{ibtehaz2023acc}           & 16.77  & 72.77             & 64.71             & 84.15             & 80.39             & 90.45             & 82.91             & 82.96 \\
\hline
UNext~\cite{valanarasu2022unext}         & 1.47   & 79.75             & \underline{70.83} & 85.73             & 83.17             & 91.10             & 79.39             & 84.26 \\
UCTransNet~\cite{wang2022uctransnet}     & 66.34  & \underline{80.31} & 71.91             & \underline{86.18} & \textbf{83.82}    & \textbf{92.23}    & \underline{82.72} & \underline{85.19} \\
\hline
PA-Net~\cite{wang2019panet}              & 8.94   & 72.22             & 68.02             & 83.76             & 80.70             & 89.34             & 80.53             & 81.83 \\
ASG-Net~\cite{li2021adaptive}            & 34.06  & 51.39             & 40.26             & 45.33             & 48.45             & 50.27             & 31.55             & 46.42 \\
UniverSeg~\cite{butoi2023universeg}      & 1.18   & 74.36             & 66.78             & 77.59             & 77.37             & 82.99             & 74.39             & 77.17 \\
\hline
MathcSeg (ours)                         & 7.79   & \textbf{80.82}    & \textbf{73.53}    & \textbf{86.44}    & 82.97             & 91.64             & 82.28             & \textbf{85.33} \\
\bottomrule 
\end{tabular}
\label{tab:nv_cross}
\end{table*}

\begin{figure*}[!htb]  
\centering  
\includegraphics[width=0.98\textwidth]{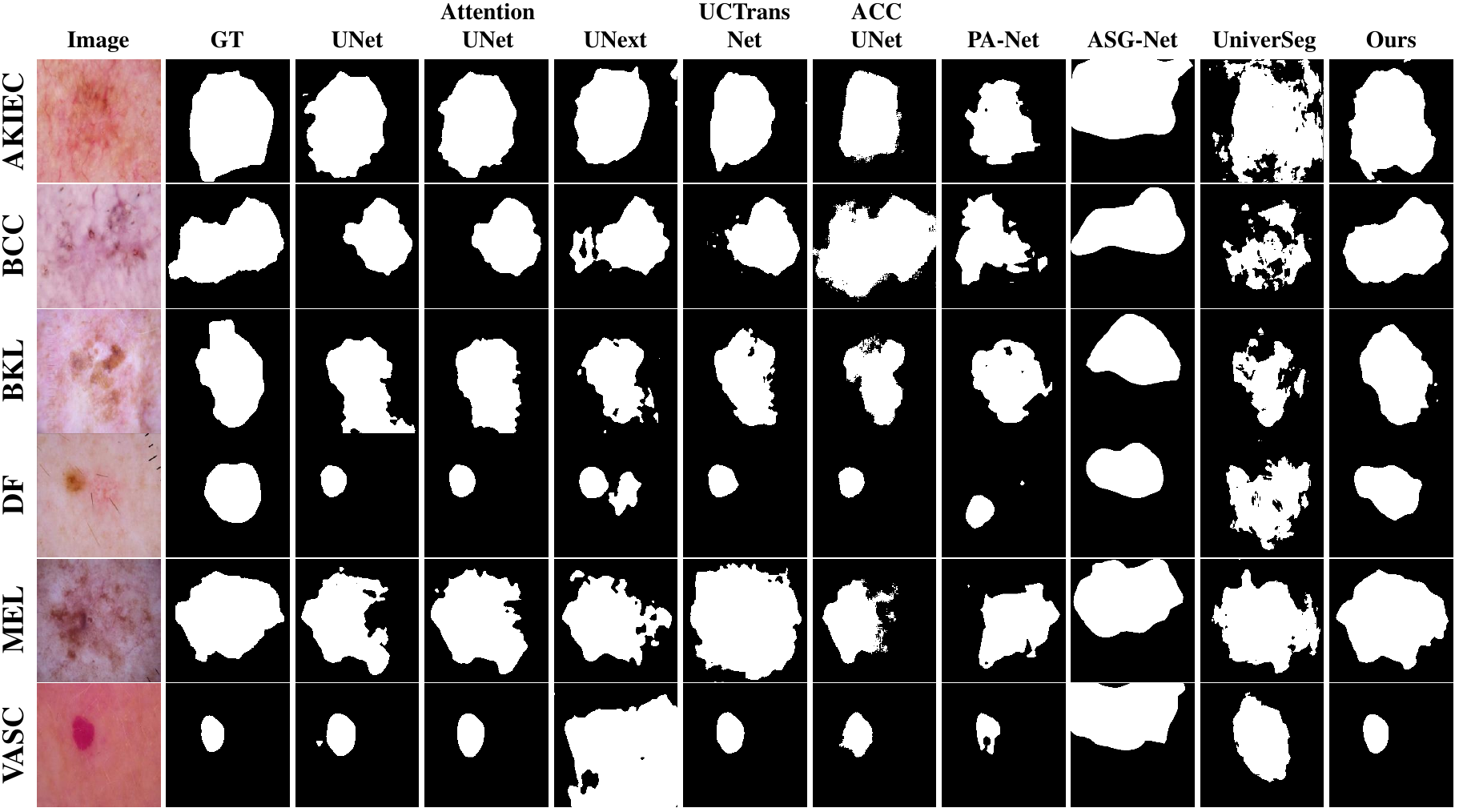}
\caption{Segmentation results visualization of different methods on the HAM10000 dataset within the cross-domain setting. All models were trained on the NV lesion type, but tested on other lesion types.}  
\label{fig:visualiztion_cross}  
\end{figure*}

\section{Experimental Results}
We compared MatchSeg to medical image segmentation models including: CNN-based models (U-Net~\cite{ronneberger2015u}, Attention U-Net~\cite{oktay2018attention} and ACC-UNet~\cite{ibtehaz2023acc}), transformer-based models (UNext~\cite{valanarasu2022unext}, UCTransNet~\cite{wang2022uctransnet}), and reference image segmentation models (ASG-Net~\cite{li2021adaptive}, PA-Net~\cite{wang2019panet} and UniverSeg~\cite{butoi2023universeg}).

\subsection{CLIP Image Encoder for Support Set Selection}
We demonstrate the effectiveness of the CLIP image encoder in selecting the support set. Specifically, we evaluate different support set sizes (8, 16, 32, 64) and perform inference on the HAM10000 dataset using the pre-trained UniverSeg model. Note that the pre-trained weights are frozen for inference. We repeated the process twenty times for each support set size to account for variations in model performance due to the stochastic nature of selecting the support set. Results are shown in Fig~\ref{fig:box_plot} blue line. The segmentation results improve with increasing support set size, consistent with~\cite{butoi2023universeg}. If all twenty predictions were combined to obtain an ensemble prediction, the segmentation performance was further improved (Fig~\ref{fig:box_plot} red line). However, when the CLIP image encoder is used for support set selection, the segmentation performance is further improved by 3\% to 4\% for each support set size (Fig~\ref{fig:box_plot} green line) over the ensemble predictions. This result indicates that selecting a support set with a similar appearance to the query image as measured in a latent space can boost segmentation performance while also reducing the computational burden of achieving the ensemble predictions.

\subsection{In-domain Setting Segmentation Results}
Table~\ref{tab:in_domain} presents the in-domain segmentation performance. MatchSeg achieves the best result across all three datasets compared to other models, which demonstrates the advantage of using the CLIP-guided support set selection method and the joint attention module. Notably, all three compared reference image segmentation methods were beaten by the CNN-based and transformer-based methods, further demonstrating the effectiveness of MatchSeg. Besides, we show the qualitative results in Fig.~\ref{fig:visualiztion_main}. From the visualization results, we find that predictions obtained from MatchSeg are more consistent with the ground truth compared to other segmentation methods.

\subsection{Cross-domain Setting Segmentation Results}
Table~\ref{tab:bus_cross} shows the cross-domain segmentation performance on BUS and BUSI datasets, and Table~\ref{tab:nv_cross} displays the cross-domain segmentation performance on the HAM10000 dataset. For both datasets, our model achieved the best result demonstrating its cross-domain generalization ability while keeping the model parameters to a relatively small number. In addition, the qualitative results are shown in Fig.~\ref{fig:visualiztion_cross} where MatchSeg gives accurate segmentation boundaries while other models give more unreliable and variable predictions.

\subsection{Ablation study}
Table \ref{tab:ablation} presents segmentation performance for an ablation study performed on three datasets to demonstrate the effectiveness of each component in MatchSeg. Only applying the CLIP image encoder to select the support set or using the Joint Attention module does not achieve the best performance, while the combination of both achieves an improved performance. This implies that support set selection and effective feature interaction are both important for the segmentation task in these datasets. 

\begin{table}[!t]  
\centering
\caption{Ablation study on BUSI, BUS and GlaS datasets.}
\fontsize{9}{11}\selectfont  
\begin{tabular}{ccccc}  
\toprule
CLIP & Joint Attention & Dataset & {DSC$\%\uparrow$} & {IOU$\%\uparrow$} \\
\hline                               
                    &                     & BUSI & 76.15          & 66.94          \\
\textbf{\checkmark} &                     & BUSI & 78.95          & 70.70          \\  
                    & \textbf{\checkmark} & BUSI & 78.96          & 70.57          \\  
\textbf{\checkmark} & \textbf{\checkmark} & BUSI & \textbf{81.03} & \textbf{72.56} \\
\hline
\hline                                             
                    &                     & BUS & 86.56          & 78.98          \\
\textbf{\checkmark} &                     & BUS & 87.99          & 81.11          \\  
                    & \textbf{\checkmark} & BUS & 89.78          & 82.16          \\  
\textbf{\checkmark} & \textbf{\checkmark} & BUS & \textbf{90.58} & \textbf{83.23} \\
\hline
\hline
                    &                     & GlaS & 87.05          & 78.14 \\
\textbf{\checkmark} &                     & GlaS & 89.37          & 81.80 \\  
                    & \textbf{\checkmark} & GlaS & 89.16          & 82.43 \\  
\textbf{\checkmark} & \textbf{\checkmark} & GlaS & \textbf{90.92} & \textbf{84.04} \\

\bottomrule
\end{tabular}
\label{tab:ablation}
\end{table}

\section{Conclusion}
In this paper, we propose MatchSeg, a framework to perform reference image segmentation in a more robust and accurate manner. MatchSeg has two key components: CLIP-guided image selection and the joint attention module to select the best support set and ensure appropriate feature alignment to the query image. Experimental results demonstrated the efficacy of MatchSeg on both domain-specific and cross-domain segmentation tasks.

\bibliography{sample.bib}
\bibliographystyle{IEEEtran}

\end{document}